\documentclass[review]{elsarticle}

\usepackage{lineno,hyperref}
\modulolinenumbers[5]

\usepackage{enumerate}
\usepackage{paralist}
\usepackage{url}
\usepackage{xcolor}
\usepackage{natbib}
\usepackage{hyperref}
\usepackage{comment}
\usepackage{multirow}
\usepackage{bm}
\usepackage{subcaption}
\usepackage{caption}
\usepackage{amssymb}
\usepackage[inline, shortlabels]{enumitem}
\usepackage[russian,english]{babel}
\usepackage{graphicx}
\graphicspath{ {./graphs/} }

\newcommand{\task}[1]{the SMM4H 2022 Task 2}

\journal{Journal of Biomedical Informatics}

\begin{document}

\begin{frontmatter}

\title{Data and models for stance and premise detection in COVID-19 tweets: insights from the Social Media Mining for Health (SMM4H) 2022 shared task}

%\tnotetext[mytitlenote]{Fully documented templates are available in the elsarticle package on %\href{http://www.ctan.org/tex-archive/macros/latex/contrib/elsarticle}{CTAN}.}

\author[sber]{Vera Davydova}
\ead{veranchos@gmail.com}
\author[UESTC]{Huabin Yang}
\ead{202121080420@std.uestc.edu.cn}
\author[kfu,hse,sber]{Elena Tutubalina}
\ead{ElVTutubalina@kpfu.ru}

\address[sber]{Sber AI, 19 Vavilova St., Moscow, Russian Federation, 117997}
\address[UESTC]{University of Electronic Science and Technology of China,  Qingshuihe Campus:No.2006, Xiyuan Ave, West Hi-Tech Zone, Chengdu, China, 611731}
%\address[airi]{Artificial Intelligence Research Institute, 5 Nizhny Susalny lane, Moscow, Russian Federation, 117997}
\address[kfu]{Kazan Federal University, 18 Kremlyovskaya street, Kazan, Russian Federation, 420008}
\address[hse]{National Research University Higher School of Economics, 11 Pokrovsky Bulvar, Moscow, Russian Federation, 109028}

% \address[spb]{St. Petersburg Department of the Steklov Mathematical Institute, 27 Fontanka, St.Petersburg, Russian Federation, 191023}
% \address[ins]{Insilico Medicine Hong Kong Ltd, Pak Shek Kok, New Territories, Hong Kong}

% \address{Radarweg 29, Amsterdam}
% \fntext[myfootnote]{Since 1880.}

% %% or include affiliations in footnotes:
% \author[mymainaddress,mysecondaryaddress]{Elsevier Inc}

% \author[mysecondaryaddress]{Global Customer Service\corref{mycorrespondingauthor}}
% \cortext[mycorrespondingauthor]{Corresponding author}

% \address[mymainaddress]{1600 John F Kennedy Boulevard, Philadelphia}
% \address[mysecondaryaddress]{360 Park Avenue South, New York}

\begin{abstract}
The COVID-19 pandemic has sparked numerous discussions on social media platforms, with users sharing their views on topics such as mask-wearing and vaccination. To facilitate the evaluation of neural models for stance detection and premise classification, we organized the Social Media Mining for Health (SMM4H) 2022 Shared Task 2. This competition utilized manually annotated posts on three COVID-19-related topics: school closures, stay-at-home orders, and wearing masks. In this paper, we extend the previous work and present newly collected data on vaccination from Twitter to assess the performance of models on a different topic. To enhance the accuracy and effectiveness of our evaluation, we employed various strategies to aggregate tweet texts with claims, including models with feature-level (early) fusion and dual-view architectures from \task{} leaderboard. Our primary objective was to create a valuable dataset and perform an extensive experimental evaluation to support future research in argument mining in the health domain.
\end{abstract}

\begin{keyword}
natural language processing \sep COVID-19 tweets  \sep opinion mining \sep argument mining \sep stance detection \sep premise detection
% \MSC[2010] 00-01\sep  99-00
\end{keyword}

\end{frontmatter}

%\linenumbers

\section{Introduction}
In recent years, social media platforms have become powerful channels for public discourse, shaping opinions, and disseminating information on various topics. The advent of the COVID-19 pandemic has further intensified the prominence of online discussions, particularly regarding critical public health measures such as masks, lockdowns, school closures, and vaccine mandates. Understanding the stances and premises expressed in these discussions is crucial for policymakers, healthcare professionals, and researchers to gauge public sentiment, identify misinformation, and develop effective communication strategies \cite{stance-detection,attitude}.

This paper addresses the challenge of automatic stance and premise detection in tweets specifically related to COVID-19 mandates. Stance detection involves determining the point of view (stance) of the text’s author towards a particular topic, while premise detection aims to identify the underlying reasons for supporting those stances.  By automatically analyzing a large volume of tweets, we can gain valuable insights into the prevailing opinions, concerns, and rationales within the online community.
Detecting stances and premises in tweets poses unique challenges due to the limited length of the messages, the informal nature of the language used, and the presence of noise and ambiguity. Furthermore, the topic of COVID-19 mandates is highly polarized, with divergent perspectives ranging from enthusiastic support to vehement opposition. Therefore, developing robust and accurate computational models capable of capturing the nuanced stance and premises expressed in tweets is vital for gaining a comprehensive understanding of the public discourse surrounding this critical issue.

A preliminary version of this work has appeared in~\cite{davydova-tutubalina-2022-smm4h,weissenbacher-etal-2022-overview}. In this journal version, we have made several significant improvements, including:
% \begin{inparaenum}[(1)]
\begin{enumerate}
\item  Annotation of an external test dataset on a new claim topic related to vaccination. This topic was not included in the training set of \task{}, which covered three other claims: school closures, stay-at-home orders, and wearing masks. The dataset is annotated by human experts, ensuring high-quality labels for training and evaluation purposes. 
\item  An extended description of the experimental datasets and emotion analysis of tweets. 
\item  An extended description of the high-scoring systems used in \task{}. These models combine deep learning algorithms and linguistic features to predict stances and identify premises within the tweet. 
\item  Investigation of model performance on different claim topics, with the addition of new experimental results and conclusions. 
\item  Error analysis of the best-performing model and a discussion of its limitations.
\end{enumerate}

% Overall, these improvements have significantly enhanced the quality and scope of our work, providing valuable insights into the challenges and new research questions of detecting claims and premises related to public health crises.

In this work, we seek to answer the following research questions: 
\begin{enumerate*}[start=1,label={\bfseries RQ\arabic*:}]
\item To what extent can models trained on specific claims be transferred to generalize and apply to other claims within the same domain?
\item Can the fusion of tweets and corresponding claims significantly enhance the performance of models?
\end{enumerate*}

% The contributions of this paper are twofold. Firstly, we introduce a curated dataset specifically tailored for stance and premise detection in COVID-19 vaccine mandates discussions on Twitter. This dataset is annotated by human experts, ensuring high-quality labels for training and evaluation purposes. Secondly, we test a few models that combine deep learning algorithms and linguistic features to predict stances and identify premises within the tweet data. 

The remainder of this paper is organized as follows. Sec. \ref{sec:rw} provides an overview of related work on stance and premise detection in social media. Sec. \ref{sec:data} describes the methodology employed in the data collection and annotation process. In Sec. \ref{sec:model}, we present the experimental setup, model architectures, and evaluation metrics. Sec. 5 describes obtained results.  Finally, in Sec. 6, we discuss the implications of our findings, limitations of the study, problems that we faced during data collection, and potential directions for future research.
All the data and code written in support of this publication are publicly available via {\texttt{\url{https://github.com/Veranchos/ArgMining_tweets}}.
% \paragraph{Stance detection}
% The first task aims to determine the point of view (stance) of the text’s author concerning the given claim (e.g., wearing a face mask). The tweets are manually annotated for stance according to three categories: in favor, against, and neither.
% \paragraph{Premise classification}
% The second task is to predict whether at least one
% premise/argument is mentioned. A given tweet is considered as having a premise if it contains a statement that can be used as an argument in a discussion. For instance, the annotator could use it to convince an opponent about the given claim. The tweets are manually annotated for binary classification: participants of this task are required to submit whether each tweet has a premise (1) or not (0).

\section{Related Work}\label{sec:rw}
\subsection{Argument Mining on COVID stances}
In recent years, there has been a growing interest in understanding public sentiment and stance on social media platforms, particularly regarding public health issues such as COVID-19. Several studies have focused on analyzing tweets to classify the stances expressed by users toward various aspects of the pandemic. This section provides an overview of the key findings and methodologies employed in previous research related to stance classification in tweets about COVID-19.

Since the beginning of the COVID-19 pandemic, new datasets and models in health-related issues have been created \cite{sakhovskiy2022multimodal,banda2021large,wuhrl-klinger-2021-claim,miao-etal-2020-twitter}. \cite{banda2021large} presented a large-scale curated dataset of over 1.12 billion tweets, growing daily, related to COVID-19 chatter. The largest dataset of Twitter users' stances in the context of the COVID-19 pandemic is COVID-CQ \cite{multu}. It consists of controversial tweets about the efficacy of hydroxychloroquine as a treatment. Similarly, \cite{wuhrl-klinger-2021-claim} presented a dataset for biomedical claim detection in Twitter posts. \cite{miao-etal-2020-twitter} created the LockdownTweets – mostly unlabelled tweet dataset, related to the lockdown policy in New York State during the pandemic. People’s opinions towards health mandates in Germany are investigated in \cite{beck-etal-2021-investigating}: first, relevant tweets were identified, and then the expressed stances were detected.
While most researchers concentrate on the stance detection task, a few datasets exist for premise classification. \cite{kotelnikov2022ruarg} gathered a collection of short texts from Russian social media, namely posts' commentaries. These data contain both statements defining the author's stance towards the given claims, and statements with premises “for” / “against” these claims.
\cite{10.1093/jamia/ocab047} also explored the premises and argumentation behind the anti-masks tweets. They analyzed 267k tweets about mask-wearing and identified the most common reasons for opposition to masks mandates using machine learning and qualitative content analysis.
% \cite{DavydovaTutubalina2022SMM4H} have done similar work for English, and presented a manually annotated corpus containing 6,156 short posts from Twitter on three topics related to the COVID-19 pandemic: school closures, stay-at-home orders, and wearing masks. This data were used in SMM4H workshop (task 2) \cite{weissenbacher-etal-2022-overview}. 
Even before the first vaccines for COVID-19 treatment were discovered, and especially with the release of vaccination programs all over the world, obligatory vaccine mandates became the central topic in social media. A few datasets in different languages with people’s reactions to vaccination orders exist at the moment: \cite{PURWITASARI2023108951} for Indonesian/Bahasa, \cite{Italian} for Italian, Arabic \cite{data7110152}. Most of the existing works in this area are dedicated to detecting the spreading of misinformation about vaccination. 
It is worth noting that while existing research has made significant contributions to stance classification in the context of COVID-19 tweets, challenges and limitations persist. Some studies have acknowledged the issue of ambiguous or conflicting stances expressed within tweets, highlighting the need for more nuanced approaches to capture the complexity of user opinions accurately.

\section{Dataset} \label{sec:data}
The dataset contains 8236 tweets and their stances and premises related to COVID-19 mandates. 
6166 tweets from this data set are designated to three claims: support or opposition to face masks, school closures, and stay-at-home orders. This collection was leveraged in our preliminary work for \task{} \cite{davydova-tutubalina-2022-smm4h}. The remaining 2070 tweets are attributed to vaccine mandates and were collected and manually annotated within this work.

The training part of the data set is based on \cite{glandt-etal-2021-stance} and consists of 3,556 tweets. There is a balanced mix of three topics: 37\%, 33\%, and 30\% of the tweets are about face masks, school closures, and stay-at-home orders, respectively. 
Both validation and test sets were collected using Twitter API. Then, the preprocessing algorithms were applied to remove tweets that were too short, contained only hashtags or mentions, or were likely considered advertisements or spam. The data collection pipeline is illustrated on Fig. \ref{fig:pipeline}.
\begin{figure}[ht!]
    \centering
    \includegraphics[width=\textwidth]{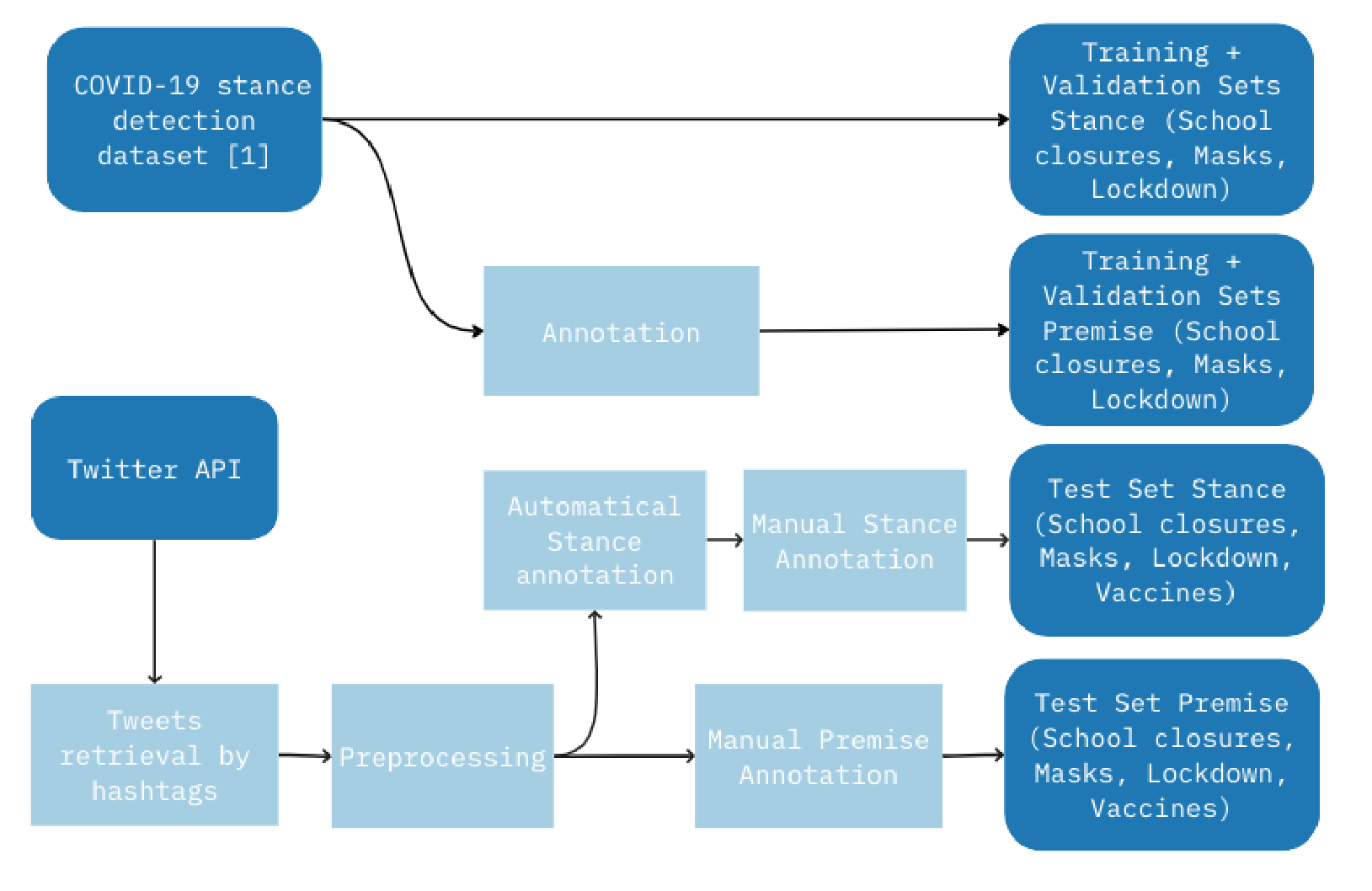}
    \caption{Data collection pipeline.}
    \label{fig:pipeline}
\end{figure}

\subsection{Vaccination tweets}
We also wanted to check whether we can leverage models trained on the data from \cite{davydova-tutubalina-2022-smm4h} to predict stances and premises related to a slightly different topic without additional training. Thus, we included tweets about an additional COVID-related mandate in the test set: vaccination. 
To collect the relevant tweets about COVID-19 vaccine mandates, we proceeded with the same approach as in our previous work related to other COVID-19 mandates. Firstly, we analyzed existing works and datasets regarding tweets about vaccine attitudes. Then, we extracted relevant hashtags and collected tweets written between January 2021 and November 2022 containing these hashtags. Since our work is dedicated to the detection of people’s stances towards COVID-related mandates, we additionally analyzed all the extracted keywords used in those tweets and allocated keywords regarding people’s stances towards vaccine mandate (i.e., \#NoMandatoryVaccines or \#vaccinatetheworld). We also manually annotated the stances of those hashtags. 
The complete list of hashtags and their stances can be found in Appendix \ref{appendix:hashtags}. Although hashtags are
great for selecting likely relevant tweets; they are
noisy and not reliable enough to accurately identify
the stance towards a target for a tweet.
Since textual data collected from social media suffers from inevitable noises, such as emoticons, URLs, and misspellings, pre-processing is performed to reduce the impact of noise. Specifically, the URLs, mentions, and hashtags were removed by using the python package \textit{tweet-preprocessor}\footnote{\url{https://pypi.org/project/tweet-preprocessor/}}, and the emoticons are converted into their textual representations by using the python package \textit{emoji}\footnote{\url{https://pypi.org/project/emoji/}}. We also removed duplicates and short tweets (less than 150 symbols). Finally, we randomly selected 3000 tweets with an equal proportion of tweets related to different automatically labeled stances (favour/against/neither). The resulting tweets were used for annotation.

\subsection{Collection and Annotation}
We followed annotation guidelines of an argument mining shared task RuArg-2022 \cite{kotelnikov2022ruarg}. Below, we highlight some of the key features of our guidelines:
\begin{itemize}
\item A statement is evaluated as an argument if it contains a statement that can be used in a dispute to persuade an opponent. For example, \textit{masks help prevent the spread of the disease.} (1)
\item It is also necessary to distinguish sentiment (positive and/or negative) from argumentation. For example, \textit{and the fact that Trump did not introduce a suffocating quarantine is well done!} (0)
\item The argument should not be a fragment that needs to be thought out. For example, \textit{It is effective if you declare a quarantine.} (0)
\item An example of an argument could be such a common sense statement. For example, \textit{in all countries of the world, everyone is wearing masks, but ours… this is not a joke.} (1)
\item The position of the author ``favor'' or ``against'' should be clear – only under this condition it is possible to detect an argument. The annotator should not think for the author. For example, the author's position on quarantine is unclear in the text \textit{here are the words of my classmate from Annecy, France, from today's Facebook correspondence - “France introduced quarantine, and immediately everyone poured out to barbecue in nature.”} (0).
\end{itemize}
Annotators had two tasks: 
1) Check whether the automatically labeled tweet’s stance is correct. If it was incorrect, they needed to change it to the correct one. 
2) Annotate whether the tweet contains a premise according to stance: label 1 means the tweet contains a premise, and label 0 means the absence of a premise. 
All tweets were annotated by five annotators, then we selected only those tweets that had a high degree of agreement between annotators (minimum 4/5 annotations were identical for each subtask). As a result, we created a new dataset containing 2070 tweets about vaccination, which was later used to test the models.
% All tweets were dually annotated; first, labels of three annotators from a crowd-source platform were aggregated into a single label \cite{10.2307/2346806}, and then the tweets were labeled by a second expert annotator. Inter-annotator agreement was 0.61 (Cohen's kappa). 
As a crowd-source platform, we use \textit{Yandex.Toloka}\footnote{\url{https://toloka.yandex.ru/}} for \task{} and 
\textit{TagMe}\footnote{\url{https://tagme.sberdevices.ru//}} for tweets about vaccination. 
Tab. \ref{tab:examples} illustrates the examples from the resulting data set. 
\begin{table*}[t!]
\centering
\scalebox{0.8}{
\begin{tabular}{|p{9.5cm}|l|l|}
\hline
\textbf{Tweet} & \textbf{Stance} & \textbf{Premise} \\
\hline
"The first and most powerful, tool we have against COVID-19 is vaccination. Vaccination is by
far the most important way to protect oursleves against severe illness, hospitalisation or death
in the event that you contract the coronavirus \#IChooseVaccination \#VaccinesSaveLives " & favor & 1 \\
\hline
100\% this man.. Vaccine passports have failed dismally in all countries who tried it. They only serve to provide a false sense of security, discrimination and segregation. \#prochoice \#NoVaccinePassportAnywhere & against & 1 \\
\hline
They're trying to do that with the fake vaccine anyway. A MANDATED vaccine for a 99.9\% survival rate is nuts. They've managed to convince the whole world they're going to die if they're not coerced into it. That's fishy. Agenda 2030 is in effect. & against & 1 \\
\hline
 Biden's vaccine mandates are the most egregious U.S. human rights violations since slavery. Biden is another Josef Mengele, the Nazi "Angel of Death." We must fight against these terrible injustices, and vote out all Democrat Party politicians. \#novaccinemandates \#impeachBiden & against & 0 \\
\hline
  It’s like  soon as the tv comes on the 1st thing I hear is either covid, vaccine or booster it’s like they have nothing else to talk about, sure enough it’s turned off within minutes	& neither & 0 \\
\hline
People infected with the omicron variant were almost 60 percent less likely to enter the hospital than those infected with delta.
Regardless odds are better with vax+ boost.  Government should be more active in advertising vaccination. & favor & 1 \\ 
\hline
Santa popped by one of our south west London vaccination centres to say thank you to all the wonderful vaccinators and volunteers! 
Get your Covid vaccination for Christmas! & favor	& 0 \\
\hline
\end{tabular}
}
\caption{Examples of tweets about COVID-19 vaccine mandates annotated for stance and premise classification.} \label{tab:examples}
\end{table*}

\subsection{Analysis of the annotated dataset}

\begin{table}[t!]
\centering
\setlength{\tabcolsep}{3pt}
\begin{tabular}{|p{2.5cm}|ccc|cc|}
\hline
\multirow{2}{*}{\textbf{Claim/Topic}} & \multicolumn{3}{c|}{\textbf{Stance}} & \multicolumn{2}{c|}{\textbf{Premise}} \\ \cline{2-6}
 & favor & against & neither & 1 & 0 \\
 \hline
\multicolumn{6}{|c|}{\task{}: train set (3566 tweets)} \\\hline
face masks & 652 & 324 & 343 & 508 & 811 \\
close school & 526 & 217 & 307 & 535 & 515 \\
home orders & 168 & 333 & 686 & 288 & 899 \\\hline
\multicolumn{6}{|c|}{\task{}: validation set  (600 tweets)} \\\hline
face masks & 121 & 51 & 36 & 82 & 126 \\
close school & 91 & 35 & 51 & 80 & 97 \\
home orders & 32 & 72 & 111 & 58 & 157 \\\hline
\multicolumn{6}{|c|}{\task{}: test set (2000 tweets)} \\\hline
face masks & 209 & 208 & 260 & 253 & 424 \\
close school & 215 & 192 & 263 & 294 & 376 \\
home orders & 102 & 170 & 381 & 169 & 484 \\\hline
\multicolumn{6}{|c|}{new annotated tweets  (2070 tweets)} \\\hline
vaccines & 421 & 279 & 1370 & 614 & 1456 \\ \hline
\end{tabular}
\caption{Summary of statistics of stance and premise classification datasets. The topic on \textit{school closures} and \textit{stay at home orders} has been shortened to \textit{close school} and \textit{home orders}, respectively.} \label{tab:stats}
\end{table}

Tab. \ref{tab:stats} shows statistics of experimental datasets across topics. The training set includes 38\% and 25\% in-favor and against tweets, respectively. Relative class balance is also present for argumentation: 63\% of train tweets contain a premise,  
34\% of tweets in the test set contain a premise; 26\% of tweets in the test set are annotated as in-favor. 
As shown in Tab. \ref{tab:stats}, the distribution of classes by topic is different. Thus, the topic of staying-at-home orders contains more “against” tweets than tweets “in-favor”, and a neutral stance is much more common for tweets about vaccination. In almost every claim, there are fewer premises in tweets. 

\subsection{Emotion Analysis}
\begin{figure}[ht!]
    \centering
    \includegraphics{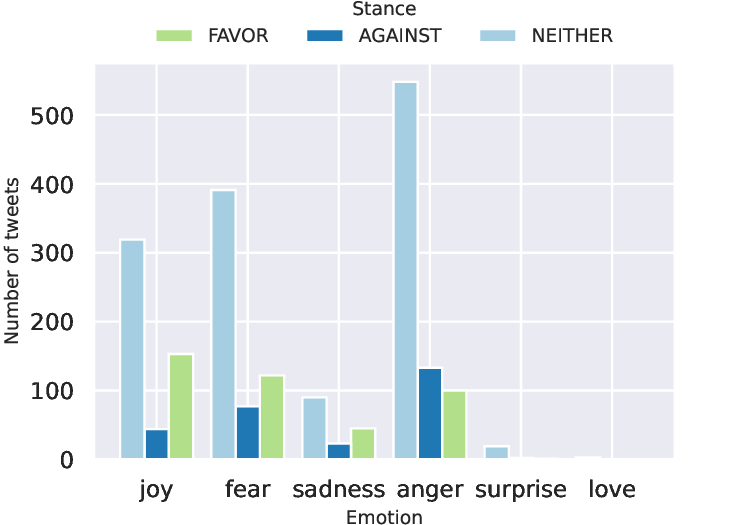}
    \caption{Distribution of 6 most common emotions in tweets per stance.}
    \label{fig:emotions}
\end{figure}

In addition to the stance and premise annotation, our research aimed to investigate the emotions expressed in tweets concerning vaccination and determine whether specific emotions correlate with the authors' positions on mandatory vaccination. We employed the emotion classification tool developed by \cite{hartmann2022emotionenglish} to address this inquiry.

The employed model, DistilRoBERTa-base, was fine-tuned on a balanced subset derived from six diverse data collections. This model predicts Ekman's six basic emotions \cite{Emotions}, along with a neutral class. Fig. \ref{fig:emotions} illustrates the distribution of emotions observed in the collected tweets. Tab. \ref{table:emotions_tweets} in the appendix illustrates some examples of tweets labeled according to the emotions expressed.

Anger emerged as the predominant emotion among all combined tweets. This observation suggests that the topic of vaccination elicits intense emotions among users. Furthermore, this finding aligns with previous research indicating that anger is a prevalent emotion in tweets pertaining to COVID-19 \cite{OLIVEIRA2023100253}. Moreover, studies have shown that tweets provoking anger are more likely to be shared \cite{anger-viral,negative-viral}. Notably, anger is predominantly expressed in tweets authored by individuals who oppose mandatory vaccination. During the annotation process, we observed that tweets adopting an opposing stance often exhibited aggressive rhetoric and insults directed towards supporters of the opposing viewpoint. While such tendencies were also observed in tweets expressing support for mandatory vaccination, they were relatively less frequent.

Another notable emotion observed in tweets with the “FAVOR” label was joy. Users frequently expressed joy regarding vaccine development, shared successful vaccination experiences, and expressed hope for the prompt resolution of the pandemic.

Furthermore, fear was found to be one of the prevalent emotions in tweets about vaccination. It is essential to mention that during our analysis, we noted instances where the model incorrectly assigned the label "fear" to more neutral tweets. We attribute this misclassification to the inherent association between the broader topic of a pandemic and the emotion of fear, potentially leading to a slight bias in the model's predictions. To mitigate this issue, we plan to incorporate manual labeling and conduct a more comprehensive analysis using data labeled by domain specialists in future research.

In summary, the most common emotions observed in tweets about mandatory vaccination are anger, fear, and joy. This finding underscores the relevance and high emotional sensitivity surrounding this topic. A deeper analysis of these emotions can contribute to the development of stance detection systems, and the insights gained can inform public opinion monitoring systems.

\section{Models}
Most teams utilized COVID-related BERT models for \task{}, along with additional techniques like regularized dropout (R-drop), to address issues with unbalanced label distribution and overfitting; see \cite{weissenbacher-etal-2022-overview} for F1 scores for each of the 14 team’s best-performing system for both tasks.

In this paper, we focus on three key model characteristics: (i) aggregation of claim and tweet texts as input data of a language model, (ii) syntactic features, and (iii) dual-view architecture that learns the representations of both types of texts.
In this regard, we employed models that proved their robustness for stance and premise detection in tweets related to COVID-19 and were ranked as the top-5 best-performing architectures during \task{}: BART + syntax features \cite{das-etal-2022-enolp} and DAN-Bert \cite{yang-etal-2022-yiriyou}. Each model leveraged different techniques to capture the nuanced information in the tweets. All the used models were trained in two set-ups: basic (utilizing the tweets only) and with the adoption of the early fusion technique \cite{early-fusion}. Early fusion employs concatenated features from two modalities (tweets texts and corresponding claims) as input features for the classifiers. The models used in our research are described below.

\subsection{BERT-base model}
Since Transformer-based models \cite{NIPS2017_3f5ee243} produce robust context-based representations of textual data, we leverage BERT-base-uncased as our base model to solve the premise and stance detection task. 
This is a BERT model consisting of 12 layers of Transformer encoder, 12 attention heads, 768 hidden size, and 110M parameters \cite{devlin2018pretraining}; we fine-tuned the model for the text classification tasks on our data. 

\subsection{BART + Syntax features} 

One of the best-performing models in \task{} leveraged the syntactic features of the text. Thus, we included this model in the comparison. This model is based on BART architecture \cite{lewis-etal-2020-bart}, an encoder-decoder transformer model with a BERT-like bidirectional encoder and auto-regressive decoder like GPT. Since BART's pre-training task encourages the model to learn representations that are robust to noise and variations in the input text, the model is well-suited for tasks that require handling noisy and ambiguous text.
To capture the syntactic structure of the input sentences, we incorporate a feature obtained from the dependency parse tree of the sentence. We leverage the open-source tool Spacy \cite{spacy2} to construct the dependency parse tree. Following that, the dependency features are encoded in descending order of their occurrence. Consider three distinct dependency tags: `aux', `amod', `nsubj' in all of the input sentences in the original dataset. If `aux', `nsubj', and `amod' have their occurrence count in the order mentioned, then `aux', `nsubj', and `amod' are encoded as one, two, and three, respectively. These encoded features are concatenated to the transformer model's output and fed into the classifier head to obtain the stance/premise. 
Instead of the usual cross-entropy loss, the supervised contrastive
loss function \cite{NEURIPS2020_d89a66c7} was used. This helped leverage label information better and morph the embedding space by pulling data points from the same class closer and pushing apart data points from other classes.
The dataset for contrastive training was created
by adding positive and negative labels to the samples present in the task dataset. We try three different strategies: fine-tuning with contrastive loss only, pre-training with contrastive loss, then fine-tuning with cross-entropy loss, and fine-tuning with a weighted loss with 0.7 weight given to cross-entropy and 0.3 to contrastive loss. The weights
were chosen considering the performance when
fine-tuning with cross-entropy and contrastive loss.

\subsection{COVID-Twitter-BERT}
BERT-large-uncased model, pre-trained on 97M unique tweets (1.2B training examples) collected between January 12 and July 5, 2020, containing at least one of the keywords ``wuhan'', ``ncov'', ``coronavirus'', ``covid'', or ``sars-cov-2''. These tweets were filtered and preprocessed to reach a final sample of 22.5M tweets (containing 40.7M sentences and 633M tokens), which were used for training \cite{muller2020covid}. We fine-tuned this model for our two classification tasks: binary classification for Premise detection and 3-class classification for Stance prediction. 
The following parameters were leveraged during this fine-tuning. Cross-entropy was leveraged as a loss function with the learning rate set to 4,00E-05. We also used AdamW \cite{loshchilov2018decoupled} as an optimizer with the weight decay set to 0. We trained the model for $10$ epochs with the batch size set to $8$ and the maximum sequence length set to 128. 

\subsection{DAN-BERT}
This architecture draws inspiration from the dual-view adaptation neural network \citep{xu2020dan}, which learns both representations of subjective and objective features of texts. Fig. \ref{fig:model} presents the overall system architecture. 

Given a tokenized tweet, the output embedding of [CLS] token is obtained through the COVID-Twitter-BERT-v2 model \cite{muller2020covid}, which is a pre-trained BERT model on a large corpus of tweets regarding COVID-19 and is the latest version with better downstream performance \footnote{\url{https://huggingface.co/digitalepidemiologylab/covid-twitter-bert-v2}}.  [CLS] is a special classification token and the last hidden state of BERT which is used for classification tasks. We employ it because of its ability to capture a representation of the whole sequence. 

\begin{figure}[t!]
    \centering
    \includegraphics[scale=0.31]{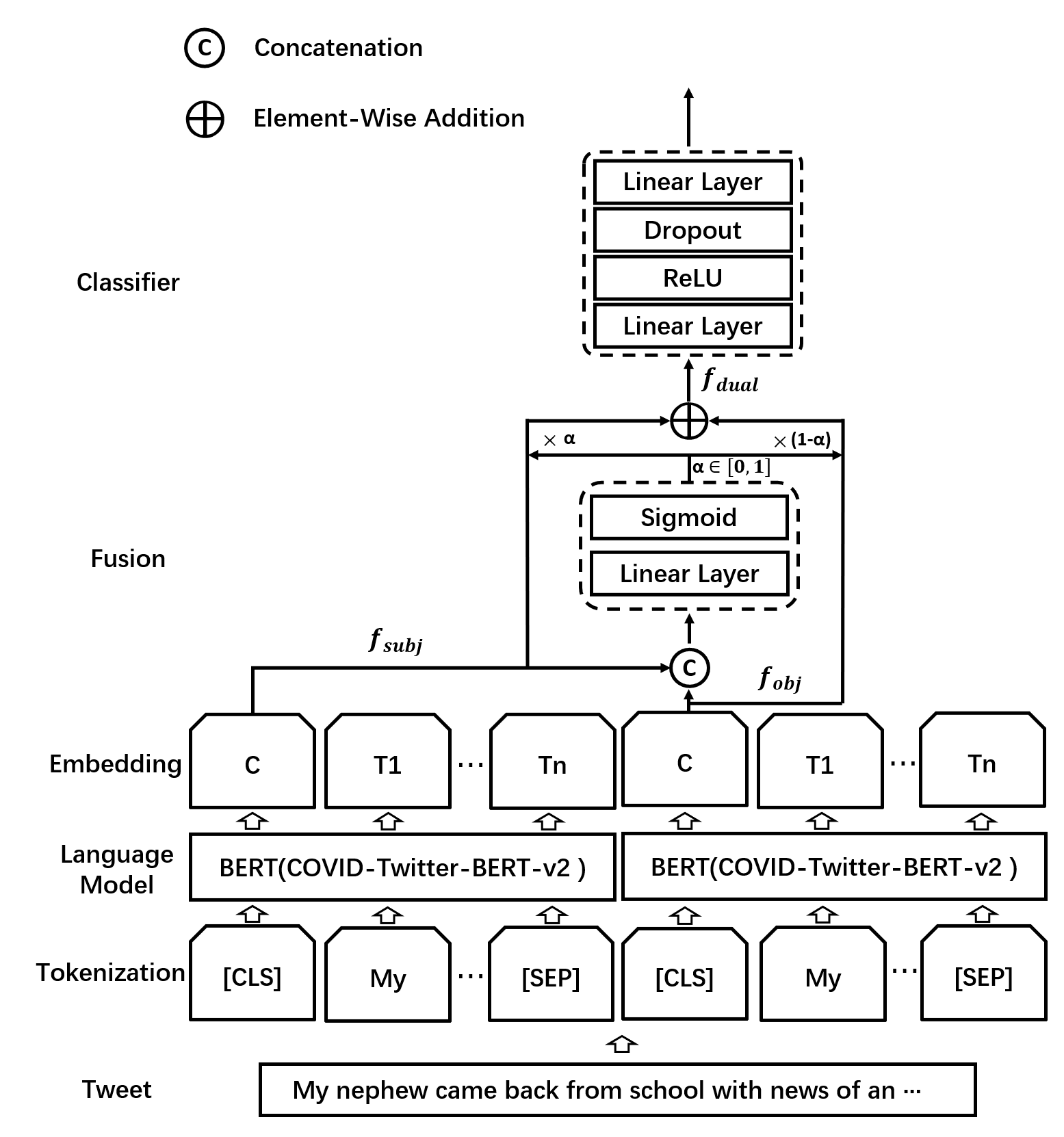}
    \caption{The overall dual-view architecture. The text is tokenized with a [CLS] token and a [SEP] token added at the beginning and end, respectively. The tokens are passed through the pre-trained language models to get their corresponding embeddings, followed by a fusion block and a classifier.}
    \label{fig:model}
\end{figure}

For extracting subjective and objective information separately, the [CLS] token is duplicated and passed through two COVID-Twitter-BERT-v2 models to obtain the corresponding embeddings (i.e., subjective and objective feature vectors, denoted by $\bm{f}_{subj}$ and $\bm{f}_{obj}$, respectively).

These two embeddings are then concatenated and fed into a fusion block composed of a fully connected layer with hidden dimensions of 2048 and a sigmoid activation function to produce a weight vector as follows,

\begin{equation}
\label{eq1}
\bm{\alpha} = {\rm sigmoid} (\bm{W}[\bm{f}_{subj}, \bm{f}_{obj}] + \bm{b}),  
\end{equation} 
where $\left[,\right]$ denotes vector concatenation and $\bm{W}, \bm{b} $ are trainable parameters. 
\par
The goal of the fusion block is to attain an optimal combination between $\bm{f}_{subj}$ and $\bm{f}_{obj}$. Once the weight vector $\bm{\alpha}$ is obtained, the weighted sum of $\bm{f}_{subj}$ and $\bm{f}_{obj}$ can be computed as follows,
\begin{equation}
\label{eq2}
\bm{f}_{dual} = \bm{\alpha} \odot \bm{f}_{subj} + (1 - \bm{\alpha}) \odot \bm{f}_{obj},
\end{equation} 
 where \(\odot\) denotes the element-wise product and $\bm{f}_{dual}$ represents the weighted sum of $\bm{f}_{subj}$ and $\bm{f}_{obj}$.
\par
Finally, the vector $\bm{f}_{dual}$ is passed through a classifier, which consists of two linear layers with hidden dimensions of 1024, a ReLU activation function, and a dropout layer with a dropout probability of 0.15.
\paragraph{Implementation}
In an implementation, there are 50 training epochs, the batch size is set to 16, the optimizer is AdamW \citep{Loshchilov2017DecoupledWD} with a learning rate of 1e-6 and a weight decay rate of 0.01, the max length of input is set to 128, and the loss function is cross-entropy loss.
\par
Additionally, cross-validation is a data resampling method used to evaluate models and prevent overfitting. The basic form of cross-validation is k-fold cross-validation \citep{refaeilzadeh2009cross}. Therefore, the 5-fold cross-validation with a majority voting strategy is applied in the implementation and also proved to improve the performance of the proposed system. 
\par
All the experiments are conducted on a server with an Inter Xeon Gold 6142M CPU and an NVIDIA GeForce RTX 3090 GPU. \label{sec:model}

\section{Results}
\subsection{Evaluation}
We leveraged F$_1$ as the main evaluation metric in each of the two subtasks, which is calculated according to the following formula: $F_1 = \frac{1}{n} \sum_{c \in C}^{} F_{1_{rel_c}}$, where $C=$\{``face masks'', ``stay at home orders'', ``school closures''\}, $n$ is the size of $C$, $F_{1_{rel_c}}$ is macro F$_1$-score averaged over two classes for each task (in-favor \& against classes for stance; 0 \& 1 classes for premise). 
The F$_1$ is also used for evaluation results obtained on vaccine mandates data set.
Systems performance for data related to school closures, masks, and stay-at-home orders (topics that were leveraged during training processes) is described in Tab. \ref{table:overall_results}. 

Separately, we evaluated the best-performing models on the data related to vaccine mandates (Tab. \ref{table:vaccine_results}). 

All the models perform significantly better than the random baseline. COVID-Twitter Bert shows the best results on the test set containing tweets about school closures, stay-at-home orders, and masks mandates. This is a large-scale model, trained specifically on the tweets about COVID-19; therefore, it shows the best results on such texts. However, this model performed second best on vaccination data. We attribute this to the fact that COVID-Twitter BERT was trained on a corpus of 160M tweets about the coronavirus collected during the period from January 12 to July 5, 2020. This time dates back to the beginning of the pandemic when vaccines had not yet been developed and corresponding mandates had not been announced.
Complex models leveraging syntax features (BART+syntax features) or dual architectures (DANBert) work much better than a general-domain language model (BERT). Their performance on the SMM4H data set is close to that of models trained on specific Twitter data (COVID-Twitter BERT). The best results on the new data set containing data about mandatory vaccination are demonstrated by DANBert.

The difference between the models' performances is statistically significant (Wilcoxon \textit{p}-value $\leq$ 0.05) \cite{Wilcoxon} on the two datasets. However, models trained on the \task{} data significantly outperform a random baseline on the vaccination dataset. It indicates that they are capable of providing useful insights and predictions on the unseen topic and can be successfully transferred to the tweets in the same domain but in slightly different theme. Moreover, adding tweet claims to the text during training can improve the performance on the unseen data in the new domain (vaccines). Nevertheless, it does not affect the results on the data in the same domains it was trained. 

\begin{table}[t!]
\centering
\begin{tabular}{*{5}{|l}|}
\hline 
\multirow{2}{*}{Model} & \multicolumn{2}{c|}{Tweets} & \multicolumn{2}{c|}{Tweets + Claims}\\
\cline{2-5} 
 &  F$_1$ Stance &  F$_1$ Premise & F$_1$ Stance & F$_1$ Premise\\
\hline
Random  & 0.268 & 0.33 & 0.268 & 0.33 \\ \hline
General-domain BERT &  0.464 &  0.352 &  0.446 & 0.354 \\ \hline 
COVID-Twitter-BERT & \textbf{0.601} & \textbf{0.719} & 0.245  & 0.68 \\ \hline
BART+Syntax & 0.45 & 0.34 & 0.38 & 0.35 \\ \hline 
DANBert  & 0.581 & 0.704 & 0.581 & 0.684 \\
\hline
\end{tabular}
\caption{\label{table:overall_results} F\textsubscript{1}-scores (F\textsubscript{1}) for stance and premise detection in tweets about school closures, masks, and lockdown}
%\end{table*}
%\begin{table*}[htbp]
\centering
\begin{tabular}{|l||l|l|l|l|}
\hline 
\multirow{2}{*}{Model} & \multicolumn{2}{c|}{Tweets} & \multicolumn{2}{c|}{Tweets + Claims}\\
\cline{2-5}  
 &  F$_1$ Stance &  F$_1$ Premise & F$_1$ Stance & F$_1$ Premise \\
\hline
Random  & 0.223 & 0.286 & 0.223 & 0.286 \\ \hline
General-domain BERT &  0.26 &  0.568 &  0.298 & 0.57 \\ \hline 
COVID-Twitter-BERT & 0.255 & 0.555 &  0.36  & 0.456 \\ \hline
DANBert  & 0.41 & 0.592 & \textbf{0.43} & \textbf{0.61} \\
\hline
\end{tabular}
\caption{\label{table:vaccine_results} F\textsubscript{1}-scores (F\textsubscript{1}) for stance and premise detection in tweets about vaccines.}
\end{table}

% \subsection{Evaluation on separated claims}
Fig. \ref{appendix:plots} presents the bar plots demonstrating the F1 metric for each model per each claim.

\section{Discussion}
\begin{table*}[t!]
\centering
\scalebox{0.8}{
\begin{tabular}{|p{9.5cm}|p{2cm}|}
\hline
\textbf{Tweet text} & \textbf{Problem}\\
\hline
\textit{Lots of cat content. Despite the vaccine, I am feeling the worst I have yet with this Breakthrough Covid case. Isolated in my room from my husbad, sons, fam for Christmas, at least I have my Edna} & unclear position\\
\hline
\textit{You really are an insensitive imbecile grow up and respect the fact that real people are hurting here leave your anti vax posts to those that care} & emotion \\
\hline
\textit{Day 645:
\#COVID19 cases keep rising! The DeltaVariant and OmicronVarient cases keeps rising! WearAMask GetVaccinated GetBoosterShot
global:
Cases: 277,167,932
Deaths: 5,377,435
US:
Cases: 51,545,991
Deaths: 812,069} & statistics \\
\hline
\textit{How many politicians do you think have these vaccine companies in their back pockets?
You don't become a millionaire by joining Congress without backdoor deals.} & consipracy theory \\
\hline
\textit{Dear sweet, innocent Marie. It's time to break you of your bubble. I'm sorry I'm the one that has to be it. \#COVIDIOTS \#antivaxxers \#coronavirus \#Unvaccinated \#GetVaccinated} & reply \\
\hline
\end{tabular}
}
\caption{Examples of tweets containing ambiguous stance.} \label{tab:problem_tweets}
\end{table*}
During the process of dataset creation, certain data characteristics emerged that posed challenges and led to disagreements among annotators. These instances also proved to be perplexing for language models (LMs). It is essential to acknowledge that these cases may hinder the accurate assessment of public opinion regarding mandatory vaccination. In this section, we will provide a brief description of these challenging cases.
Firstly, we encountered tweets that strongly endorsed vaccines; however, the author's stance on mandatory vaccination remained ambiguous. Tab. \ref{tab:problem_tweets} presents examples illustrating such cases.
Additionally, a considerable number of tweets consisted of news reports or statistical information. Despite implementing algorithms to filter out such texts, they still appeared in the datasets since they were not solely posted by news accounts but also by real users.
Moreover, certain tweets implicitly conveyed the author's position, requiring additional contextual information for accurate stance assessment. For instance, the last row in Tab. \ref{tab:problem_tweets}: a tweet reacting to another user's remark without access to the original post can challenge determining the stance expressed.
Furthermore, emotional tweets were prevalent, which could be perplexing as the emotion was evident, but the stance was not explicitly presented. The provided example (second row in Tab. \ref{tab:problem_tweets}) showcases clear aggression, allowing us to infer the author's support for vaccination. However, the tweet does not address the attitude towards mandatory vaccination.
Additionally, a substantial number of tweets expressed support for various conspiracy theories about vaccination. However, even in these cases, the author's stance on COVID-19 mandates is not always apparent.
Researchers should consider that the aforementioned cases could potentially impede a comprehensive understanding and accurate assessment of public opinion regarding mandatory vaccination.
\subsection{Potential applications}
Understanding public sentiment toward healthcare policies, institutions, or specific healthcare issues can be crucial for healthcare providers and policymakers. Stance detection models can help gauge public opinion and adjust strategies accordingly. Another potential application of such systems is disinformation identification. Social media is rife with healthcare misinformation and disinformation. Determination and categorization of content that spreads false or misleading information about healthcare can help authorities respond promptly with accurate information or warnings. In conclusion, the presented data set about vaccine mandates aims to promote research in tracking discussions around vaccines and identify instances of vaccine hesitancy. Healthcare organizations can use this information to design targeted vaccine education campaigns to address specific concerns and increase vaccination rates.

\section{Conclusion}
The objective of the study was to evaluate various models for stance detection and premise classification on COVID-related tweets. We observed a strong interest in \task{}, with 47 participants registered, and 14 teams submitted their prediction for both tasks. The models included classifiers on general-domain LMs and LMs, pre-trained specifically on COVID-related tweets, models leveraging syntax features, and dual architectures. We found that models trained on \task{} train set of three claims could be successfully transferred to tweets in a new COVID-related claim. As mentioned above, it would be interesting to conduct a more in-depth analysis to study the relationship between emotions and stances on biomedical topics.

\section*{Acknowledgements}
The work has been supported by the Russian Science Foundation grant \# 23-11-00358. 

\bibliography{main}

\section*{Appendix}

\appendix

\begin{table*}[t!]
\label{table:emotions_tweets}
\centering

\begin{tabular}{|p{9.5cm}|p{2cm}|}
\hline
\textbf{Tweet text} & \textbf{Emotion}\\
\hline
\textit{Santa popped by one of our south west London vaccination centres to say thank you to all the wonderful vaccinators and volunteers! Make yourself a present by getting your Covid vaccination for Christmas!} & joy\\
\hline
\textit{Believe in God! Protect yourself from these satan worshippers injecting this poison into you! This is not a vaccine, but Trojan horse to introduce nanotechnology which will be integrated with travel, Commerce and digital currency etc. They want to kill billions by 2025!} & fear \\
\hline
\textit{They've already begun in Quebec "Vaccination" of the 5-11yr olds started around December 6th  (date varied on schools/sectors) These sick! Just before Christmas too..Absolutely disgusting! \#LeaveOurKidsAlone  \#NoMandatoryVaccines} & anger \\
\hline
\textit{Very caring and sensible of you. Our Xmas is cancelled too, as I'm high risk for both Covid and the vax, tho thankfully managed to get my 1st, in hospital earlier this week, without too many issues. But not risking my being out about over the Christmas break. Stay safe over there, darling.} & love \\
\hline
\textit{Everyone is tired and exhausted. It’s made much worse when you don’t see the light at the end of the tunnel. For so long that vaccination rate was the light. We’d be ok when we got there, now what? What’s keeping the hope alive now? The next booster? } & sadness \\
\hline
\textit{Wow, I'm shocked! Maybe they should reconsider the vax mandates. } & surprise \\
\hline
\end{tabular}
\caption{Examples of tweets about COVID-19 vaccines and corresponding emotions.} \label{table:emotions_tweets}
\end{table*}

\begin{figure}[ht!]
    \centering
    \includegraphics[width=\textwidth]{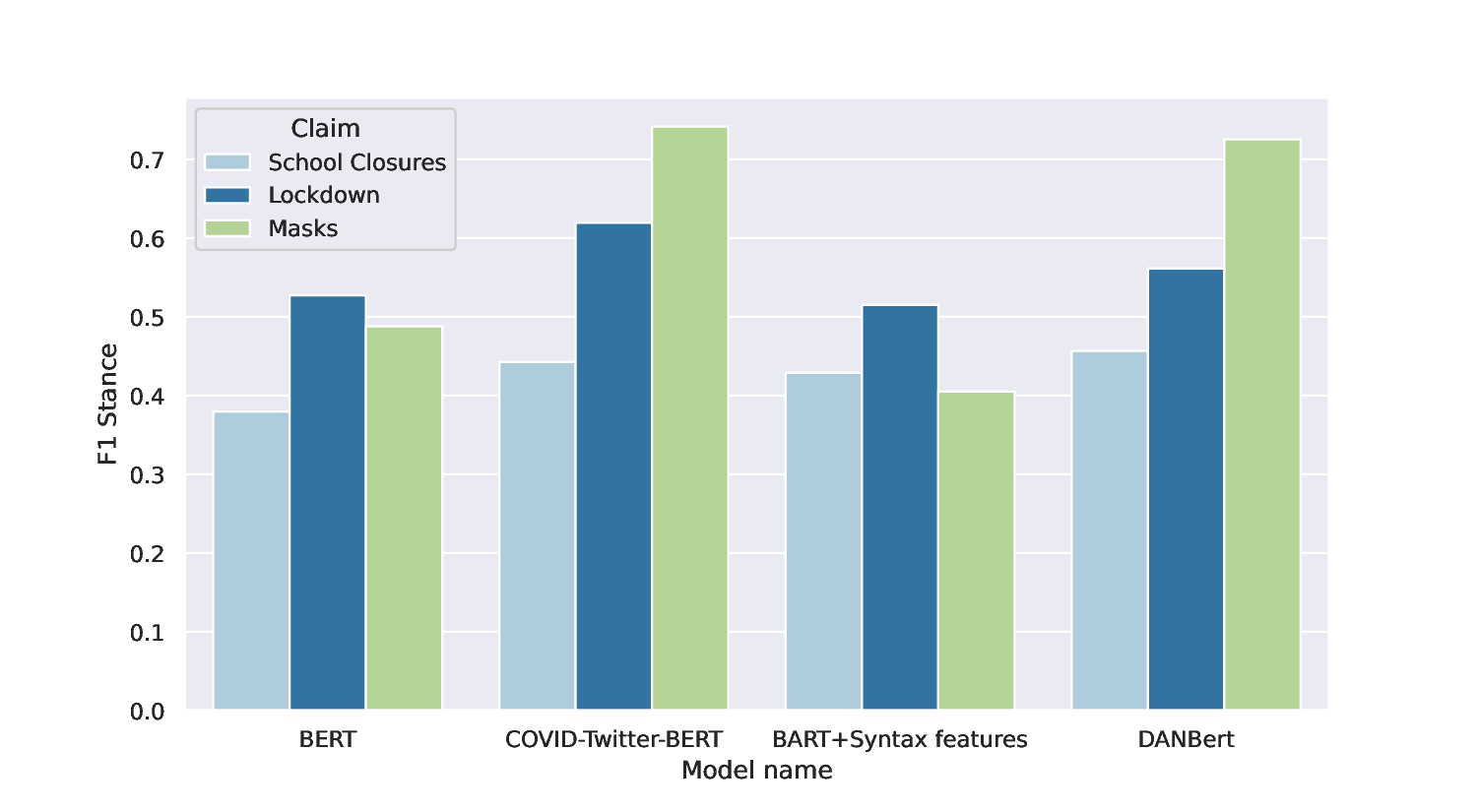}
    \caption{F1 metrics for stance detection on 3 claims: School closures, Stay-at-home orders. All models were trained on tweets about these claims.}
    \label{appendix:plots}
\end{figure}

\begin{figure}[ht!]
    \centering
    \includegraphics[width=\textwidth]{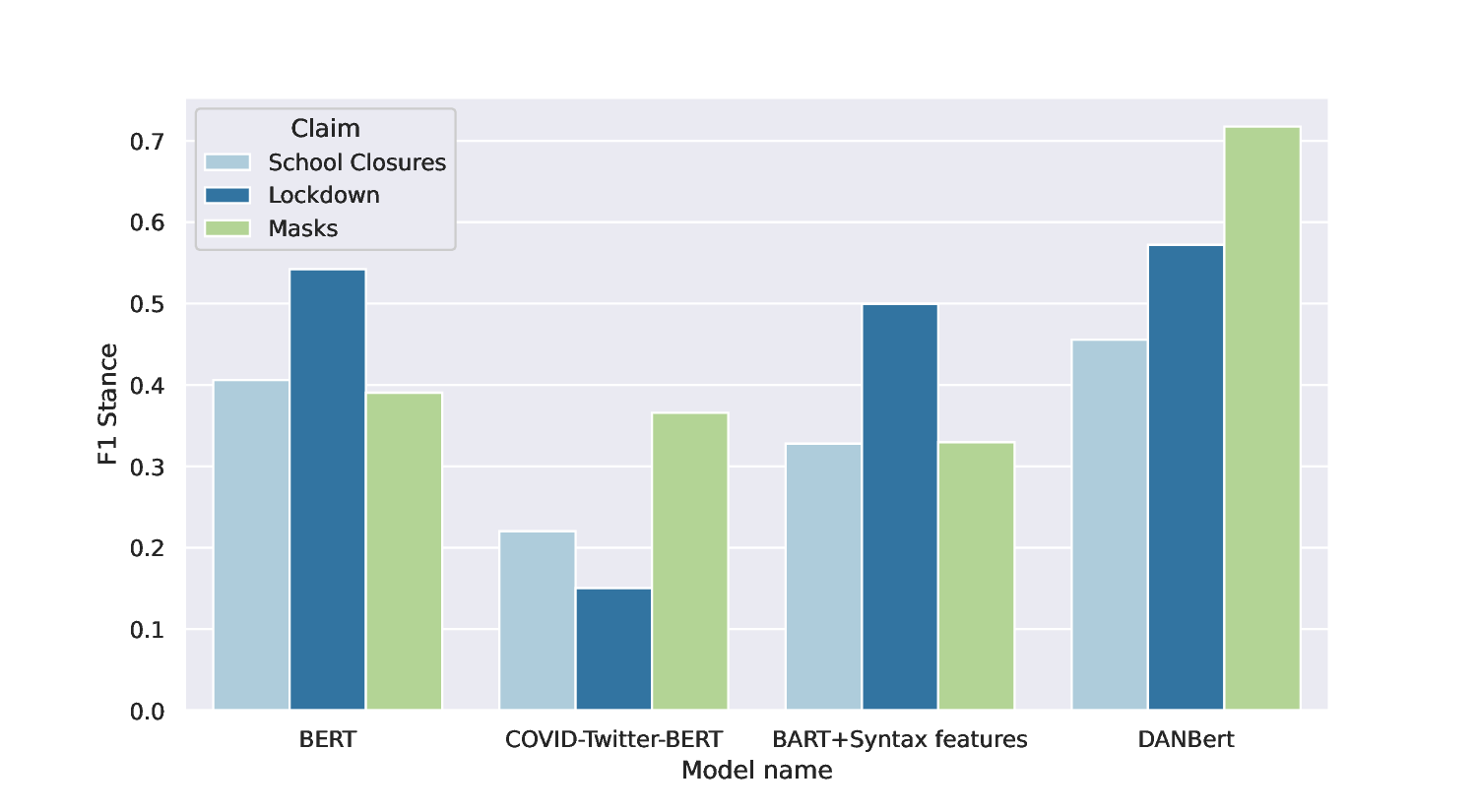}
    \caption{F1 metrics for stance detection on 3 claims: School closures, Stay-at-home orders. All models were trained on tweets and corresponding claims.}

\end{figure}

\begin{figure}[ht!]
    \centering
    \includegraphics[width=\textwidth]{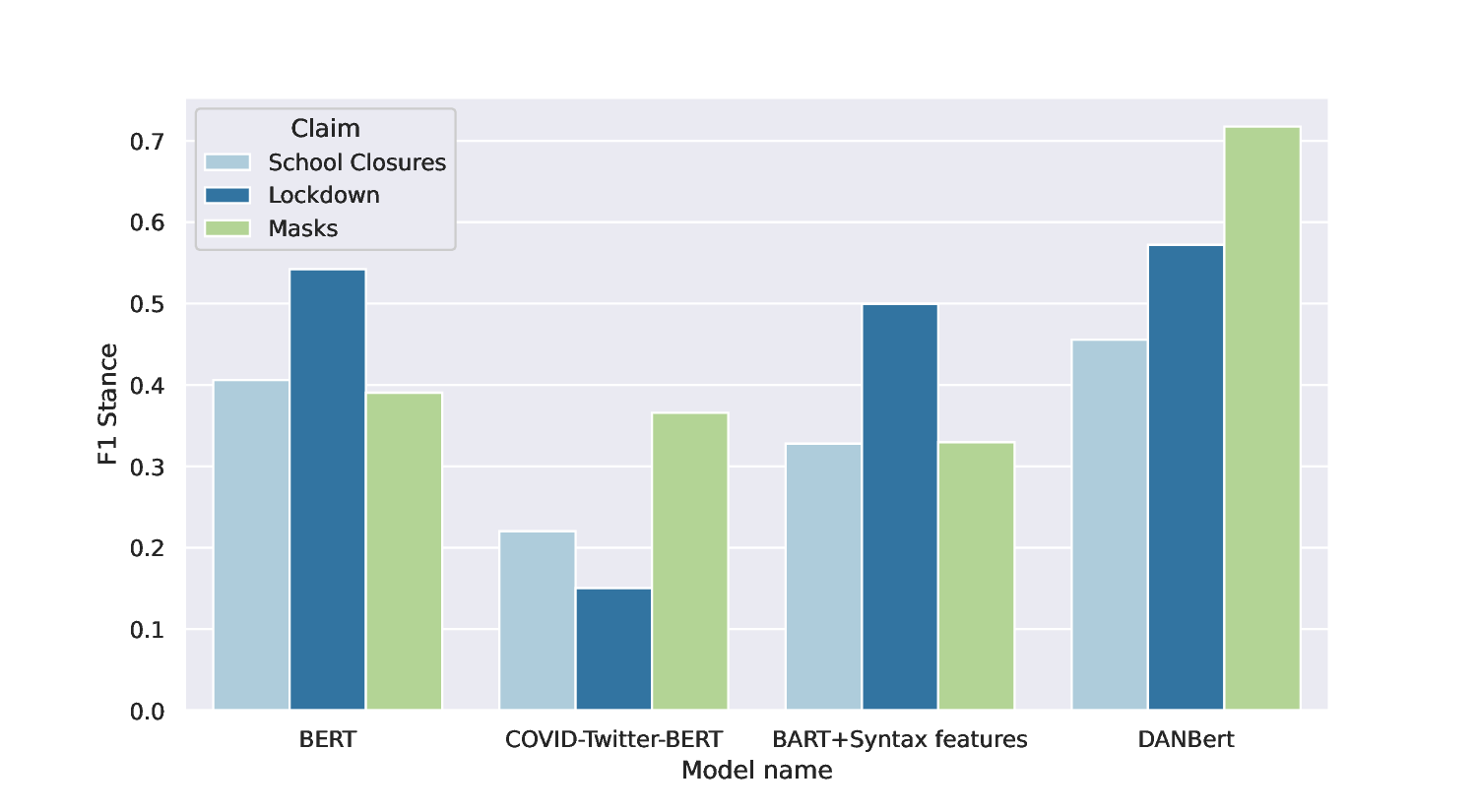}
    \caption{F1 metrics for premise detection on 3 claims: School closures, Stay-at-home orders. All models were trained on tweets about these claims.}

\end{figure}

\begin{figure}[ht!]
    \centering
    \includegraphics[width=\textwidth]{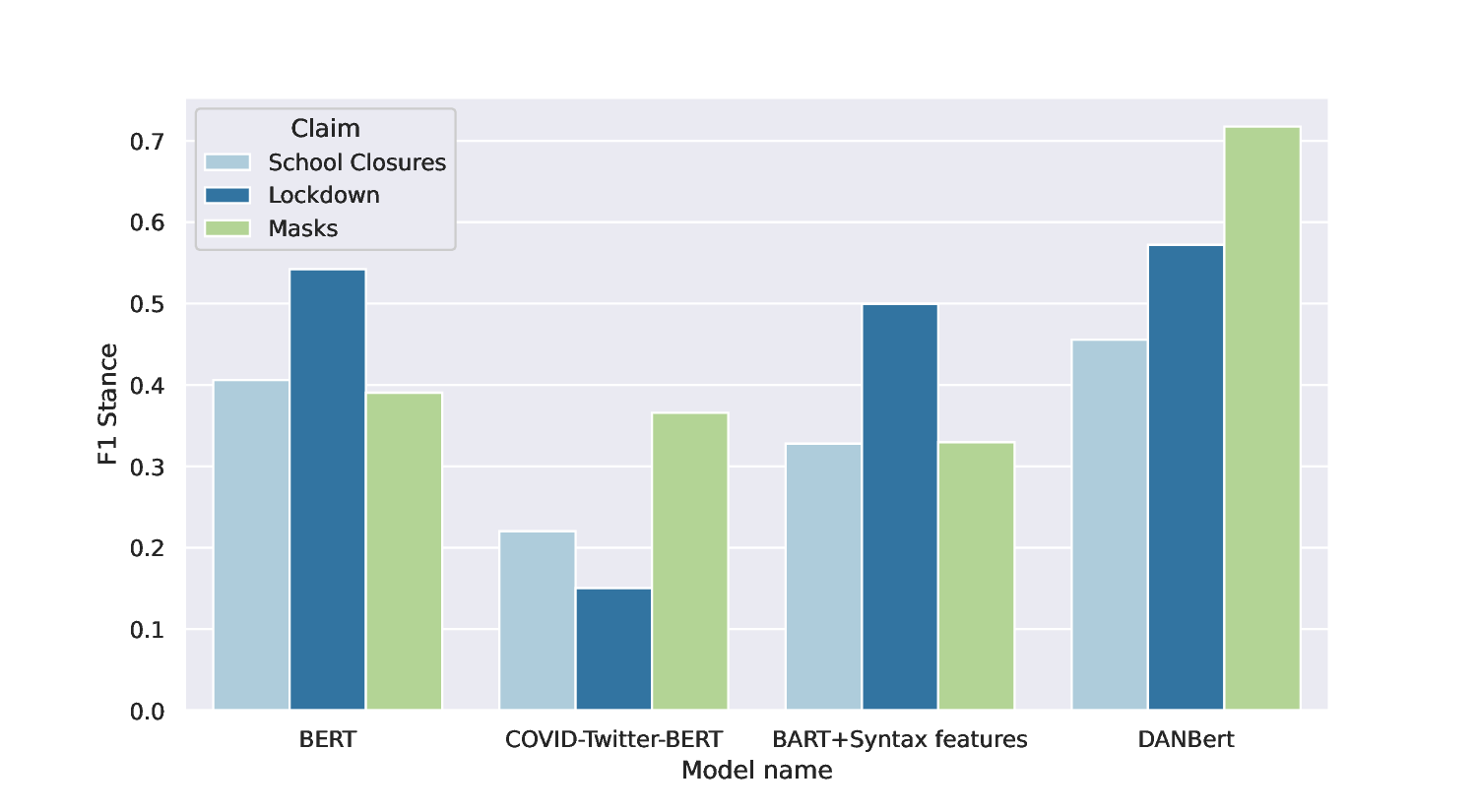}
    \caption{F1 metrics for premise detection on 3 claims: School closures, Stay-at-home orders. All models were trained on tweets and corresponding claims.}

\end{figure}

\begin{table*}[ht]
\label{appendix:hashtags}
\centering
\begin{tabular}{|p{10.5cm}|l|}
\hline
\textbf{Hashtags} & \textbf{Stance}\\
\hline
GetVaxxed
VaccinationWorks
getvaccinated
VaxxedSquad
MoreVaccinesMoreLivesSaved
vaccinesafetyadvocate
justwantmyjab
IChooseVaccination
VaccinesSaveLives
vaccineworks
getvaccined
YourBodyYourChoice
GetVaxed
GetVaccinatedX3
vaccinesavelifes
GETVACCINATED
VaccineMandatesNOW
vaccineswork
vaccinatetheworld
GetVaccinatedASAP
Vaccine4All
VaccinateTheWorld
GetVaccinatedNow
vaccinateASAP
vaccinateyourself
VaccinesWork
getvaccinatedtobeprotected
Vaccinate4All
GetVax
vaccinessaveslives
iChooseVaccination
VaxToTheMax
GetVaccinated & favor \\
\hline
NoVaccine
novaccinemandates
Vaccinationisachoice
NoVaccineMandates
NoVaxPass
NoVaccineMandate
vaccinedeaths
vaccineinjured
UNVACCINATED
NoVaccinePassportsAnywhere
VaccineDeaths
mybodymychoice
vaccineinjury
pharmacide
NoVaccinePassports
EndVaccineMandatesNow
C19vaxKills
NoVax
NoVaccinePassport
NoMandatoryVaccines
NoVaccinePassportAnywhere
vaccineinjuries
NoVaccine\_NoPandemic
jabskill
NoVaccineMandatesAnywhere
saynotoVaccinemandate
prochoice
NoVaccinePassportAnywhere & against \\
\hline
\end{tabular}
\caption{List of hashtags related to vaccine mandates used for tweets extraction and pre-annotation of tweets. Individual hashtags are separated by spaces.} \label{appendix:hashtags}
\end{table*}

\end{document}